\documentclass[a4paper]{article}

\usepackage{todonotes}

\usepackage{INTERSPEECH2019}
\usepackage{arabtex}
\usepackage{utf8}
\setcode{utf8}

\usepackage{subcaption}
\usepackage{booktabs}
\usepackage{url}

\usepackage{numprint}
\usepackage{tipa}
\usepackage{enumitem}

\title{Analyzing Phonetic and Graphemic Representations \\ in End-to-End Automatic Speech Recognition  
}
\name{ Yonatan Belinkov$^{1,2}$, Ahmed Ali$^3$, James Glass$^1$}
\address{
  $^1$MIT Computer Science and Artificial Intelligence Laboratory, Cambridge, MA, USA \\ 
  $^2$Harvard John A.\ Paulson School of Engineering and Applied Sciences, Cambridge, MA, USA \\ 
  $^3$Qatar Computing Research Institute, HBKU, Doha, Qatar
  }
\email{\{belinkov,glass\}@mit.edu, amali@qf.org.qa}

\begin{document}

\maketitle
\begin{abstract}
End-to-end neural network systems for automatic speech recognition (ASR) are trained from acoustic features to text transcriptions. In contrast to modular ASR systems, which contain separately-trained components for acoustic modeling, pronunciation lexicon, and language modeling, the end-to-end paradigm is both conceptually simpler and has the potential benefit of training the entire system on the end task. However, such neural network models are more opaque: it is not clear how to interpret the role of different parts of the network and what information it learns during training. 
In this paper, we analyze the learned internal representations in an end-to-end ASR model. We evaluate the representation quality in terms of several classification tasks, comparing phonemes and graphemes, as well as different articulatory features. We study two languages (English and Arabic) and three datasets, finding remarkable consistency in how different properties are represented in different layers of the deep neural network. 

\end{abstract}

\section{Introduction}

Traditional automatic speech recognition (ASR) systems employ a modular design, with different modules for acoustic modeling, pronunciation lexicon, and language modeling, which are trained separately. In contrast,  end-to-end (E2E) models are trained to convert acoustic features to text transcriptions directly, potentially optimizing all parts for the end task. Unfortunately, they are also less interpretible: identifying what different parts do and what 
properties they capture is less straightforward.  

It is a common problem in many neural network models besides E2E ASR. Therefore, a line of work is concerned with deciphering the information captured by learned representations in neural models that are trained on some downstream task~\cite{belinkov2019analysis}. 
Previous work analyzed different neural representations and various properties, such as evaluating how phonetic information is captured in neural acoustic models~\cite{nagamine2015exploring,Nagamine2016}. 
However, E2E ASR models are still relatively under-explored. 

In previous work~\cite{belinkov:2017:nips}, we anlyzed DeepSpeech2~\cite{amodei2016deep} E2E models, 
from the perspective of the phonetic information that is learned in different layers. However, that work only considered TIMIT as a source of phonetic information. In this paper, we extend this analysis to multiple languages (English and Arabic) and three different datasets, as well as explore additional properties (e.g., phonemes vs.\ graphemes). 
We find that over many different configurations---languages, datasets, linguistic properties---the E2E models exhibit strikingly similar behavior across layers. We also investigate the drop in representation quality at the top layers, attributing part of it to the focus on graphemes and long-distance information. 

\textbf{Limitations:} 
Potential limitations are the restrictions to a specific E2E architecture and to frame classification. 
Future work can explore other architectures and  larger segments.

\section{Related Work}
\subsection{Analysis of Representations}

Several studies  analyzed what phonetic information is encoded in acoustic models using  clustering and classification methods to~\cite{nagamine2015exploring,Nagamine2016}.
Others correlated the behavior of gates in recurrent neural networks with phoneme boundaries~\cite{wang2017gate,wu2016investigating} or visualized skip connections in speech enhancement models~\cite{santos2018investigating}. 
Various phonetic and speaker features were investigated in speaker embeddings~\cite{shon2018frame,Wang2017}, and properties like style and accent were 
analyzed in a convolutioanl ASR performance prediction model~\cite{elloumi2018analyzing}. 
Another line of work is concerned with developing and 
analyzing joint audio-visual models~\cite{chrupala2017representations,K17-1037,harwath2017learning,Drexler2017AnalysisOA}. 

Recent work~\cite{krug2018neuron} clustered neurons in  convolutional E2E ASR  and found that lower layers  encode phonemes better than graphemes. 
Most related,  our previous investigation of E2E ASR~\cite{belinkov:2017:nips}  used the same E2E model and analyzed phoneme representations only on English in TIMIT. In contrast, here we explore two different languages (English and Arabic) and three datasets. We also consider new aspects such as different phonetic features and representing past and future information.   

\subsection{E2E and Arabic ASR} \label{sub:e2e_asr}

Recently, E2E ASR has attracted attention in both academia and industry. The E2E system is based on a single deep neural network that can be trained from scratch to directly transcribe speech into  labels (words, phonemes, etc.)~\cite{amodei2016deep,miao2015eesen,pratap2018w2l}. It integrates disjoint modules, developed from traditional hybrid methods, into one network. While  attention-based models~\cite{wang2018stream,chan2016listen} address the ASR problem as sequence mapping  using an encoder-decoder architecture, the connectionist temporal classification (CTC)~\cite{graves2006connectionist, miao2015eesen} objective function performs frame-level classification with specialized time-aggregation. 

Previous work made various attempts to reduce word error rate (WER) in Arabic ASR on the MGB-2 dataset~\cite{ali2016mgb}. While initial work used phoneme-based systems~\cite{ali2014complete}, recent work, and the wining submissions, are grapheme-based~\cite{khurana2016qcri, smit2017aalto}. These efforts 
reduced WERs from $32\%$ to $14\%$. 
Since most systems focused on traditional ASR with separate acoustic, pronunciation, and language models, Arabic E2E is  still unexplored. More broadly,  it is important to understand the difference between grapheme and phoneme modeling in  E2E setups.

\section{Methods}
\subsection{Analysis by Classification}

To analyze the representation quality in the E2E ASR model, we adopt the paradigm of classification or probing tasks~\cite{belinkov2019analysis,adi2016fine,hupkes2017visualisation}. 
First, we train the E2E model in the normal fashion, on pairs of utterances and transcriptions.  Then, we run the trained model on a dataset with frame-level annotations, such as phoneme labels, and record activations from different layers of the E2E model. 
These activations are fed to a classifier that is trained to predict the labels. A separate classifier is trained for every annotation type (say, phonemes or graphemes) and layer. 

To maintain consistency with our previous analysis of DeepSpeech2~\cite{belinkov:2017:nips}, the classifier is a simple feed-forward neural network with one hidden layer of size $500$. The input and output sizes are determined by the feature representation from the E2E model and by the label set size, respectively. 
It is non-contextual, taking only the current frame representation, although context may be captured in the representation itself via the ASR model. 
The classifier is trained for 30 epochs and the model with the best validation loss is used for evaluation. 

The code for running our experiments is publicly available.\footnote{\url{http://github.com/boknilev/asr-repr-analysis}}

\subsection{Classification Tasks}

We consider the following classification tasks:
\begin{itemize}[leftmargin=0.55cm]
    \item \textbf{Phonemes}: for every frame, predict an aligned phoneme.  
    \item \textbf{Graphemes}: for every frame, predict an aligned grapheme. 
    \item \textbf{Phonetic features}: for every frame, predict the  place or  manner of articulation of its aligned phoneme.  
\end{itemize}

\subsection{Obtaining Labels} \label{sec:asr}

Our objective here is to estimate the correct timing of a sequence of phonemes for a speech signal given verbatim transcription. We use triphone HMM models with speaker information similar to~\cite{hosom2002automatic}. It is worth noting that timings from the Viterbi-alignment results are not as precise as the manually-aligned TIMIT data. Therefore, we consider running phoneme classification using a bigger window to overcome potential shift in the timing as shown in Section~\ref{res:diff_datasets}.
We also note that although  using forced-aligned phonemes is a possible limitation, the experimental results show consistent patterns with our previous analysis using manual segmentation from TIMIT~\cite{belinkov:2017:nips}.

\begin{table}[t]
\centering
\caption{Statistics of annotated datasets.}
\begin{subtable}[b]{\linewidth}
\caption{Number and total duration of utterances.} 
\centering
\begin{tabular}{l rrr rrr }
\toprule
& \multicolumn{3}{c}{Utterances} & \multicolumn{3}{c}{Hours:Minutes} \\ 
\cmidrule(lr){2-4} \cmidrule(lr){5-7}
& Train & Dev & Test & Train & Dev & Test \\ 
\midrule
Librispeech  & $1920$ & $241$ & $240$ & 3:40 & 0:26 & 0:30 \\ 
TedLIUM & $345$ & $79$ & $76$ & 1:05 & 0:15 & 0:14 \\ 
MGB-2 & $1990$ & $288$ & $295$ & 3:12 & 0:31 & 0:33 \\
\bottomrule
\end{tabular}
\end{subtable}
\begin{subtable}[b]{\linewidth}
\caption{Label set sizes.} 
\centering
\begin{tabular}{l r r r r }
\toprule
& Phonemes & Graphemes & Place & Manner  \\ 
\midrule
English & $40$ & $28$ & $9$ & $7$ \\ 
Arabic  & $34$ & $37$ & $12$ & $9$ \\ 
\bottomrule
\end{tabular}
\end{subtable}
\label{tab:annotated-data}
\end{table}

\section{Experimental Setup}

\subsection{E2E ASR Systems} 

We use standard, large-scale datasets for training the E2E ASR models. 
For English, we use a pre-trained reimplementation~\cite{Naren2016} of DeepSpeech2~\cite{amodei2016deep}. This model has $2$ convolutional layers and $5$ recurrent long short-term memory (LSTM) layers, trained with CTC. It was trained on Librispeech~\cite{panayotov2015librispeech}.\footnote{This model is referred to as DeepSpeech2-light~\cite{Naren2016}, and we used it instead of the deeper DeepSpeech2 model because we found it to work better in the Arabic case. Note that the two architectures behave similarly qualitatively according to previous work~\cite{belinkov:2017:nips}.}

For Arabic, we train our own model using the implementation in \cite{Naren2016}, with the same architecture. 
We use the MGB-2 dataset~\cite{ali2016mgb}, which comprises 1,200 hours of broadcast videos from the Aljazeera Arabic TV channel.  
We exclude sentences longer than 30 seconds and sentences failed to align with the seed models trained in~\cite{khurana2016qcri}. This give us a total training data of 700 hours, which consists of more than 312K sentences. 

\subsection{Modular ASR Systems}

To obtain phoneme and grapheme labels, we employed traditional ASR systems with separate components for phoneme or grapheme modeling. This allows us to run forced-alignment and get annotated data to be used by the classifier. 

For English, we train time-delay neural network acoustic-models using the implementation in~\cite{peddinti2015time} on the standard TedLIUM corpus~\cite{rousseau2012ted}, comprising more than $210$ hours and $92$K sentences. For the phoneme system, we use the official pronunciation
dictionary that has more than $152$K words and  $40$ phonemes/monophones. Meanwhile, the English grapheme-based lexicon is formed from the $26$ alphabet letters /a-z/, and was constructed similar to the system described in~\cite{wang2018phonetic}.

For Arabic, we train the same architecture as the English system on the MGB-2 corpus using the implementation in~\cite {khurana2016qcri}. The phoneme system used the phonetic lexicon shared in the MGB-2 challenge~\cite{ali2016mgb}, while the grapheme lexicon used the same word list with 1:1 word-to-character mapping to keep the same vocabulary size. Both lexicons have more than $950$K words. For a detailed comparison of phoneme vs.\ grapheme Arabic ASR, see Section 7.2.3 in~\cite{ali2018multi}.

\subsection{Classification Data}

Given the modular ASR systems, we annotated a subset of each dataset with forced-aliged phonemes and graphemes. 
For articulatory features, we mapped phonemes to place and manner of articulation.\footnote{We used 
TIMIT and Wikipedia (\texttt{English\_phonology}) for English mappings and  another  mapping for Arabic: 
\url{http://sites.middlebury.edu/arabiclingusitics2014/files/2014/02/class6_phonetics_1.pdf}. When classifying by 
place, we set the 
vowels as a separate group.} 
We used TedLIUM and Librispeech for English, and \mbox{MGB-2} for Arabic. 
We made sure to train and evaluate the classifier on a portion of the data not used for training the ASR models. 
Table~\ref{tab:annotated-data} provides statistics on the annotated datasets.

\begin{figure}[t]
\centering
\includegraphics[width=0.9\linewidth]{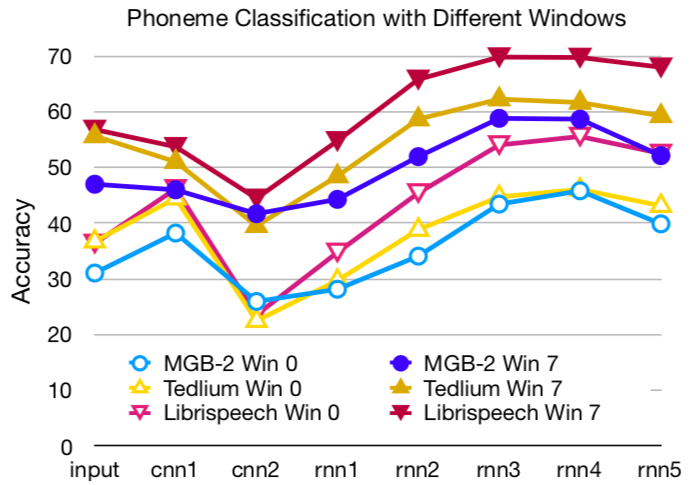}
\caption{Phoneme classification accuracy in different datasets. 
}
\label{fig:window}
\end{figure}

\begin{figure}[t]
\begin{subfigure}{\linewidth}
\includegraphics[width=\linewidth]{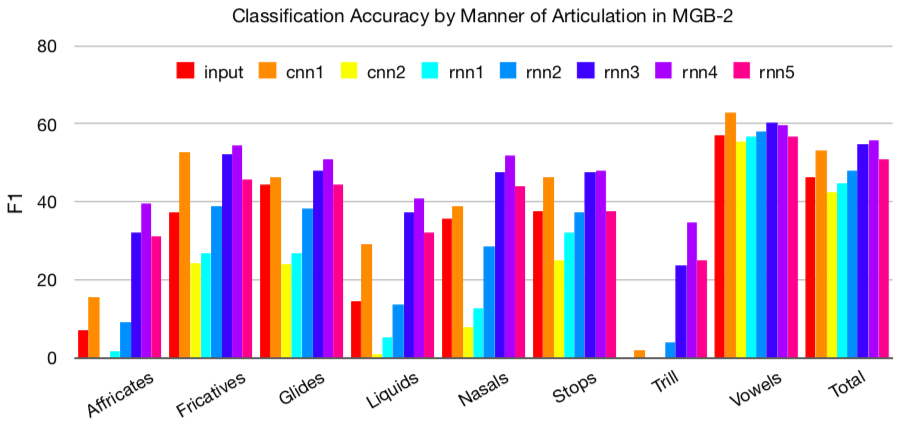}
\caption{MGB-2 (Arabic).}
\end{subfigure} \\ 
\begin{subfigure}{\linewidth}
\includegraphics[width=\linewidth]{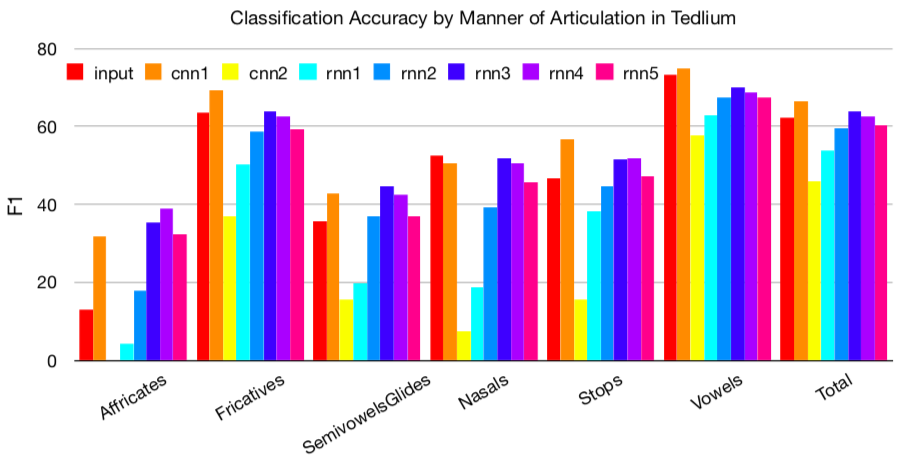}
\caption{TedLIUM (English).}
\end{subfigure}
\caption{Classification by manner of articulation.}
\label{fig:manner}
\end{figure}

\section{Results} 

\subsection{Results in Different Datasets} \label{res:diff_datasets}

Figure~\ref{fig:window} shows the result of phoneme classification in three datasets: The Arabic MGB-2 and English TedLIUM and Librispeech. The overall layer-wise trend is similar in all cases: the first convolutional layer improves representation quality above the input spectrograms, while the second convolutional layer leads to a large drop. In the  LSTM layers, there is steady increase until the last layer, where performance drops. This pattern is consistent with our previous analysis of phoneme representations in DeepSpeech2 based on TIMIT classification \cite{belinkov:2017:nips}.

We also compare classification using only the current frame vs.\ using a window of +/-$7$ frames around it. As Figure~\ref{fig:window} shows, while using additional context always helps performance on the classification tasks, the layer-wise patterns do not change, consistent with~\cite{belinkov:2017:nips}. 
Interestingly, throughout the recurrent layers the difference between using a window or not becomes smaller. For instance, on Librispeech, the difference is around $20\%$ at layers rnn1--2, decreasing to $14$--$15\%$ at layers rnn3--5; a similar pattern is found in the other datasets. 
This indicates that the top recurrent layers capture more context, thereby reducing the benefit from a large context at the input.

\subsection{Phonetic Features} 
In this section, we analyze the representation quality from the perspective of articulatory features. We map each phoneme to either its place or its manner of articulation.

Figure~\ref{fig:manner} shows the manner classification results, summarized by F1 score (harmonic mean of precision and recall) for each manner of articulation. 
In most cases, the common layer-wise pattern recurs. Some manners are easier to classify than others:
especially vowels, which are very different from consonants, and also fricatives, nasals, and stops. Affricates are more difficult, perhaps due to their composite nature. The Arabic liquid (/l/) and trill (/r/) are also hard. 
Comparing English and Arabic, the results are fairly consistent, as shown by the similar shape of the two sub-plots, although the labels do not entirely overlap. To test this quantitatively, Table~\ref{tab:corr} (right) shows the Pearson correlation across layers between several manners of articulation in English and Arabic. In most clases, there are high positive correlations, up to $0.97$ in the case of English semivowels/glides and Arabic glides. Vowels are less correlated, which is not surprising given the limited vowel inventory in Arabic (only $3$ vowels) compared to English.

\begin{figure}[t]
\begin{subfigure}{\linewidth}
\includegraphics[width=\linewidth]{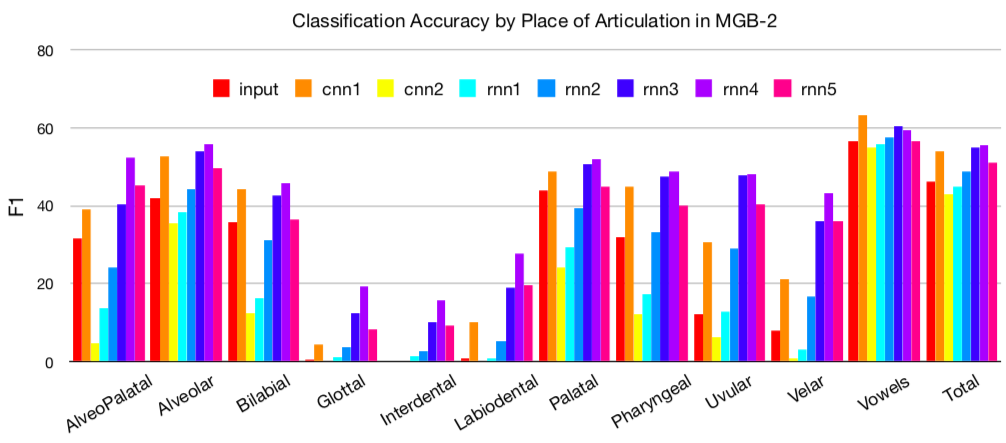}
\caption{MGB-2 (Arabic).}
\end{subfigure} \\ 
\begin{subfigure}{\linewidth}
\includegraphics[width=\linewidth]{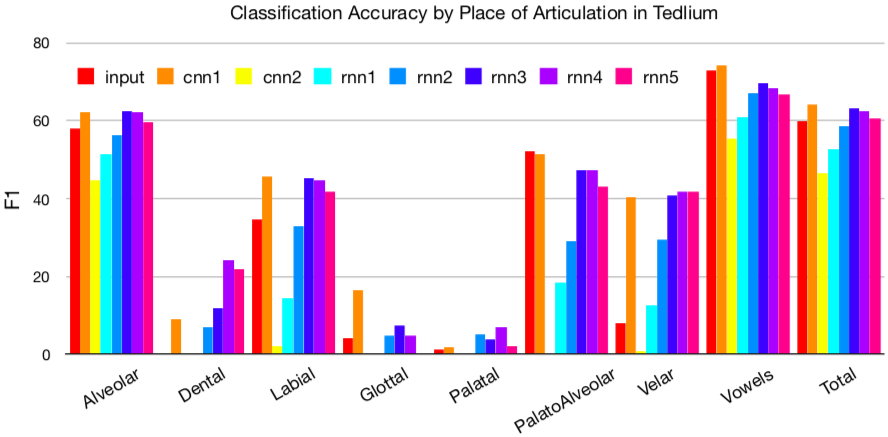}
\caption{TedLIUM (English).}
\end{subfigure}
\caption{Classification by place of articulation.}
\label{fig:place}
\end{figure}

\begin{table}[t!]
\centering
\caption{Cross-language correlation in layer-wise classification accuracy by  manner and place of articulation.} 
\npdecimalsign{.}
\nprounddigits{2}
\begin{tabular}{p{1cm}p{1cm} n{1}{2} | p{1.3cm}p{1cm} n{1}{2} }
\toprule
\multicolumn{3}{c}{Place}  & \multicolumn{3}{c}{Manner}   \\ 
TedLIUM & MGB-2 & \multicolumn{1}{c}{$r$} & TedLIUM & MGB-2 & \multicolumn{1}{c}{$r$}  \\ 
\midrule 
Glottal & Glottal & 0.159055307 & Vowels & Vowels & 0.735209577  \\
Palatal & Palatal & 0.677036977 & Fricatives & Fricatives & 0.847477359  \\
Vowels & Vowels & 0.713236949 & Semi/Glides & Liquids & 0.878873597  \\
Labial & Labiodent. & 0.731066884 & Stops & Stops & 0.911426748  \\
Palato-Alveolar & Alveo-Palatal & 0.880850767 & Nasals & Nasals & 0.92650614  \\
Dental & Interdent. & 0.893260332 & Affricates & Affricates & 0.930780778  \\
Velar & Velar & 0.908134526 & Semi/Glides & Glides & 0.96667709  \\
\cmidrule(lr){4-6}
Alveolar & Alveolar & 0.926772941 & Total & Total & 0.818908365  \\
Labial & Bilabial & 0.978108495 &  &  &   \\
\cmidrule(lr){1-3}
Total & Total & 0.8741995639 &  &  & \\
\bottomrule 
\end{tabular}
\label{tab:corr}
\end{table}

\begin{figure*}[t]
\begin{subfigure}{.3\linewidth}
\includegraphics[width=\linewidth]{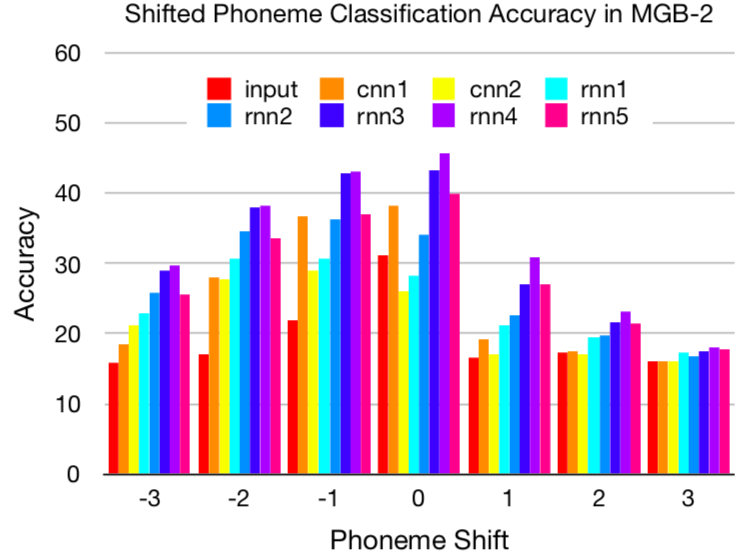}
\caption{MGB-2, phonemes (Arabic).}
\label{fig:shift-mgb2-phoneme}
\end{subfigure}
\begin{subfigure}{.3\linewidth}
\includegraphics[width=\linewidth]{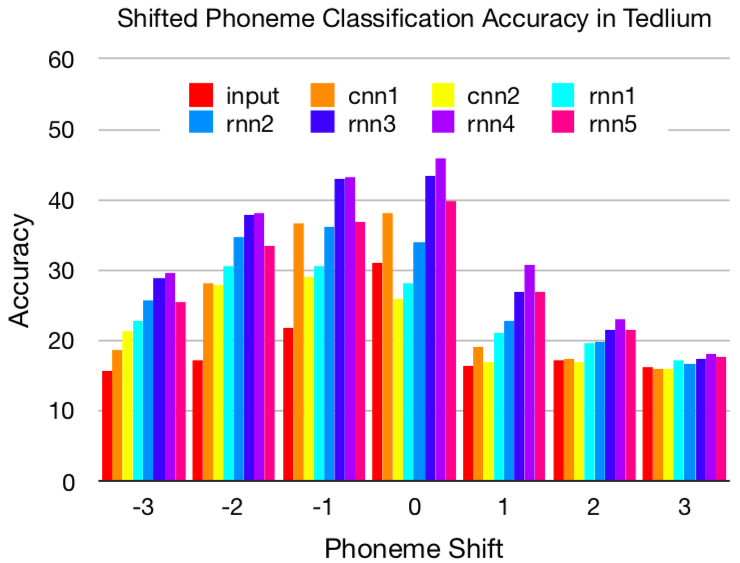}
\caption{TedLIUM, phonemes (English).}
\label{fig:shift-tedlium-phoneme}
\end{subfigure}
\begin{subfigure}{.3\linewidth}
\includegraphics[width=\linewidth]{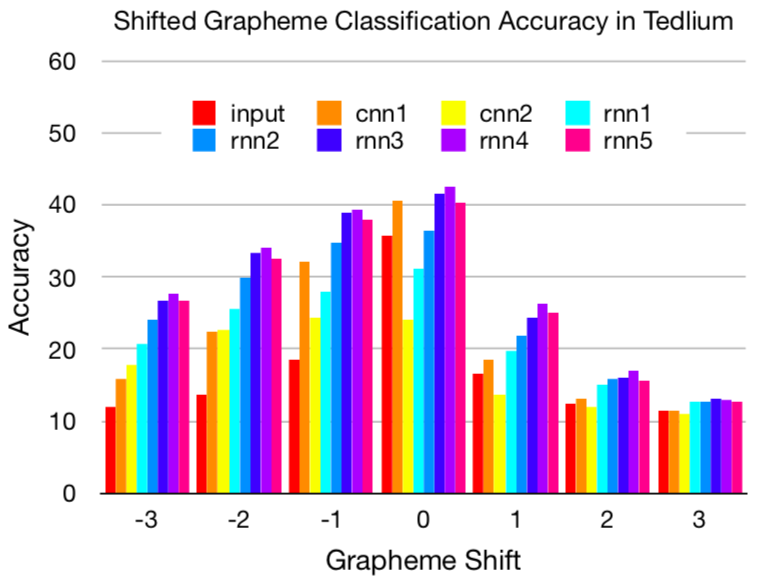}
\caption{TedLIUM, graphemes (English).}
\label{fig:shift-tedlium-grapheme}
\end{subfigure}
\caption{The effect of predicting past and future phonemes and graphemes.}
\label{fig:shift}
\end{figure*}

Turning to place of articulation, Figure~\ref{fig:place} exhibit similar layer-wise patterns in classifying each place. Some places are easier to classify: 
alveolar, alveo-palatal, labial, and velar consonants. Glottal and dental/interdental consonants are more difficult. 
Again, these results are quite consistent for the two languages, although the place labels  also do not entirely overlap. Looking at the correlations (Table~\ref{tab:corr}, left), several consonant groups behave very similarly in the two languages: labial/bilabial ($r=0.98$), alveolar ($0.93$), velar ($0.91$), and dental/interdental ($0.89$). 
This is striking as these groups do not always overlap: English labials include /b/, /p/, /v/, /f/, and /m/, while Arabic bilabials include /b/, /m/, and /w/, yet their correlation is very high. 
Cases of low(er) correlations are the glottal ($0.16$), palatal ($0.68$), and labial/labiodental consonants ($0.73)$. In the glottal case, this may be explained by Arabic having glottals /\textglotstop/ 
and /h/, while the English phoneme set only has /h/.

\subsection{Phonemes vs.\ Graphemes}

How can we explain the drop in representation quality towards the top layers of the model? 
One possibility is that a model that was trained on acoustics-to-characters ``forgets'' some of the phonetic information at the top layer(s), close to the output. To test this, we performed several grapheme classification tasks.

\begin{figure}[t]
\centering
\includegraphics[width=\linewidth]{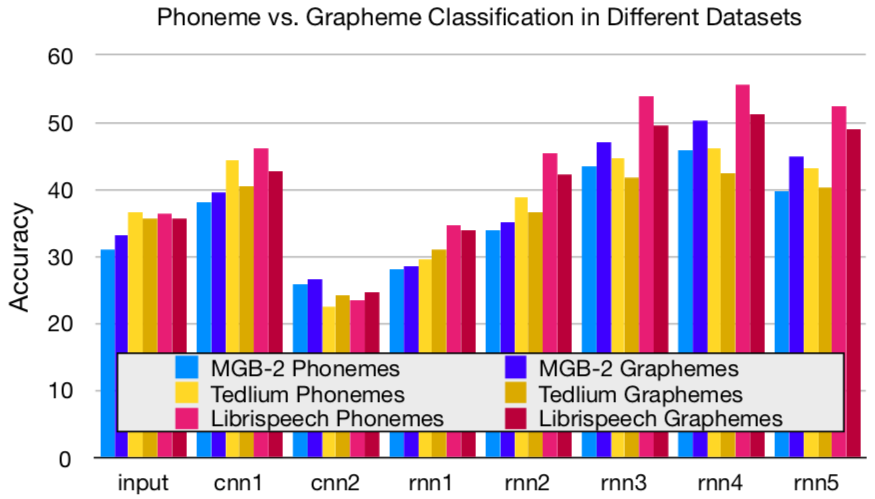}
\caption{Phoneme and grapheme classification accuracy.}
\label{fig:graphemes}
\end{figure}

Figure~\ref{fig:graphemes} shows the results. 
Evidently, the layer-wise patterns are very similar to the phoneme case, although grapheme classification tends to be slightly easier. Interestingly, the gaps between grapheme and phoneme classification are somewhat larger at the top recurrent layers than in intermediate layers. This suggests that the top layers are indeed more geared towards graphemic than phonetic information. However, the drop at the top layer is still apparent and cannot be explained solely by phoneme/grapheme differences. 
Table~\ref{tab:drops} compares the top layer drop in in phoneme and grapheme classification. In all three datasets, the (relative) drop is smaller when predicting graphemes than phonemes. This again indicates that the top layers are more concerned with graphemes than with phonemes. 
\begin{table}[h]
\centering
\caption{Relative drop from the penultimate to ultimate layer in phoneme vs.\ grapheme classification.}
\npdecimalsign{.}
\nprounddigits{2}
\begin{tabular}{l n{2}{2} n{2}{2} n{2}{2} } 
\toprule
& \text{TedLIUM} & \text{Librispeech} & \text{MGB-2} \\ 
\midrule
Phonemes & $6.358996634\%$ & $5.49193316\%$ & $12.93950644\%$ \\ 
Graphemes & $5.045515489\%$ & $4.345793482\%$ & $10.60229958\%$ \\ 
\bottomrule
\label{tab:drops}
\end{tabular} 
\end{table}

\subsection{Predicting the Future or Past} 

Another possible explanation for the drop at the top layer has to do with capturing long-distance information. 
Previous work \cite{belinkov:2017:nips} conjectured that ``higher layers in the model are more sensitive to long distance information that is needed for the speech recognition task, whereas the local information that is needed for classifying phones is better captured in lower layers.'' 
We investigate this conjecture by predicting past or future phonemes. We simply shift the labels in the datasets by $k \in \{-3,-2,-1,0,1,2,3\}$ phonemes, and retrain the classifier. The results are shown in Figures~\ref{fig:shift-mgb2-phoneme} and~\ref{fig:shift-tedlium-phoneme}.

We find that predicting the future is much more difficult than predicting the past, as performance quickly drops when predicting even only one phoneme into the future, but only moderately degrades when predicting up to three phonemes into the past. This can be explained by the use of unidirectional LSTM layers in the models we experiment with. 
This holds in both languages (compare Figures \ref{fig:shift-mgb2-phoneme} and \ref{fig:shift-tedlium-phoneme}) and in both phoneme and grapheme classification (compare Figures \ref{fig:shift-tedlium-phoneme} and \ref{fig:shift-tedlium-grapheme}). 

The drop in accuracy at the top layer is still apparent. 
In the case of Arabic phonemes (Figure~\ref{fig:shift-mgb2-phoneme}), 
this drop  is more moderate when predicting future phonemes: in relative terms, we see a drop of $2$--$6\%$ when predicting $2$ or $3$ phonemes into the future, but $14\%$ drop when predicting into the past. 
In English phonemes (Figure~\ref{fig:shift-tedlium-phoneme}), there is only a very mild drop ($1.4\%$) when predicting $3$ phonemes into the future. 
Thus, the top layer may be losing less long-distance information about the future than the past. This is not always consistent, however, as there is a substantial drop ($12\%$) when predicting $2$ phonemes into the future in English. 
In the case of graphemes, the top layer drop is fairly consistent in all shifts: predicting $3$ graphemes into the future or past results in a similar drop of around $3\%$. 

\section{Conclusion}

In this work, we analyzed an E2E speech recognition model in terms of phonetic and graphemic representations. We observed consistent behavior in layer-wise quality across languages, datasets, output labels, and articulatory features. This suggests that such models may benefit from sharing information, for example using multilingual systems as in a recent E2E codeswitching ASR model~\cite{luo2018towards}. 
In the future, we plan to extend the analysis to other E2E models, such as attentional sequence-to-sequence~\cite{wang2018stream,chan2016listen} or purely convolutional models~\cite{pratap2018w2l}.

\bibliographystyle{IEEEtran}

\bibliography{mybib}

\end{document}